\documentclass[conference]{IEEEtran}

\usepackage{etoolbox}
\makeatletter
\patchcmd{\@makecaption}
  {\scshape}
  {}
  {}
  {}
\makeatletter
\patchcmd{\@makecaption}
  {\\}
  {.\ }
  {}
  {}
\makeatother

\usepackage{verbatim}
\usepackage{cite}
\usepackage{amsmath,amssymb,amsfonts}
\usepackage{graphicx}
\usepackage{multirow}
\usepackage{textcomp}
\usepackage{xcolor}
\usepackage{mathtools}
\usepackage{float}
\usepackage[utf8]{inputenc}
\usepackage{xparse}
\ifCLASSOPTIONcompsoc \usepackage[caption=false,font=normalsize,labelfont=sf,textfont=sf]{subfig}
\else
\usepackage[caption=false,font=footnotesize]{subfig}
\fi

\usepackage[ruled,lined, noend,linesnumbered]{algorithm2e}
\usepackage{algorithmicx}
\usepackage{hyperref}

\usepackage{titlesec}
\titlespacing*{\section}
{0pt}{1.5ex}{1.5ex}
\titlespacing*{\subsection}
{0pt}{1.5ex}{1.5ex}

\def\BibTeX{{\rm B\kern-.05em{\sc i\kern-.025em b}\kern-.08em
    T\kern-.1667em\lower.7ex\hbox{E}\kern-.125emX}}
    
\graphicspath{{./}}
    
\begin{document}

	\title{\textbf{On Using Large-Batches in Federated Learning}}

	\author{\IEEEauthorblockN{Sahil Tyagi}
		\IEEEauthorblockA{\textit{Indiana University Bloomington} \\
		\href{https://orcid.org/0009-0007-8314-4745}{\textit{ORCID}}
	}}
	
	
	\maketitle
	
	\begin{abstract}
		Efficient Federated learning (FL) is crucial for training deep networks over devices with limited compute resources and bounded networks.
With the advent of big data, devices either generate or collect multimodal data to train either generic or local-context aware networks, particularly when data privacy and locality is vital.
FL algorithms generally trade-off between parallel and statistical performance, improving model quality at the cost of higher communication frequency, or  vice versa.
Under frequent synchronization settings, FL over a large cluster of devices may perform more work per-training iteration by processing a larger global batch-size, thus attaining considerable training speedup.
However, this may result in poor test performance (i.e., low test loss or accuracy) due to generalization degradation issues associated with large-batch training.
To address these challenges with large-batches, this work proposes our vision of exploiting the trade-offs between small and large-batch training, and explore new directions to enjoy both the parallel scaling of large-batches and good generalizability of small-batch training.
For the same number of iterations, we observe that our proposed large-batch training technique attains about 32.33\% and 3.74\% higher test accuracy than small-batch training in ResNet50 and VGG11 models respectively.
	\end{abstract}
	
	
	\section{Introduction}

\textit{Collaborative} or \textit{Federated learning (FL)} methods are optimized to perform on-device training when clients are resource-constrained \cite{b22, b23}, communication latency and bandwidth is bounded \cite{b3}, and data privacy or locality is paramount \cite{b0, b24}.
These constraints arise from the explosive growth in big data and Internet-of-Things (IoT), as personal edge or mobile devices generate and collect data across various modalities, presenting new opportunities device-local, context-aware deep networks.
For e.g., edge devices like NVIDIA AGX/Nano accelerate training/inference at the edge, while hardware-software co-design like Apple Intelligence runs contextual foundation models on proprietary phones and tablets \cite{b26}.
FL with readily available resources over a large number of clients with varying data distributions present various challenges pertaining to training performance and speedup.

Different federated algorithms offer distinct trade-offs between parallel and statistical performance.
Distributed training can be either \emph{scaled-up} by batching more samples per-iteration, or \emph{scaled-out} by adding more clients to collaboratively train a globally shared model.
Parallel performance or scaling is thus affected by factors like training topology, network latency and bandwidth, number of participating clients, batch-size, model-size and communication collective \cite{b3}.
The statistical performance of deep neural networks (DNNs) is influenced by its communication frequency, training hyperparameters (like learning-rate, batch-size, etc.) and data distribution, i.e., if training data across clients is independent and identically distributed (i.i.d.), unbalanced or skewed.
As we see in \S \ref{sec:relatedwork}, synchronization or aggregation frequency directly affects convergence quality; FL with higher communication volume produces models with higher accuracy (especially under non-i.i.d. settings), while synchronous training \cite{b8} attains the highest accuracy \cite{b22, b1}.
However, synchronous training over a large number of clients may result in excessively large global batches that may attain high training accuracy, but low test accuracy/poor generalization \cite{b9, b11, b17}.

Batch-size is a vital training parameter that affects both parallel and statistical performance of DNN training \cite{b15, b5}.
Batching more samples on a client can leverage parallelism at an intra-device level (i.e., scaling-up), while inter-device aggregation of updates increases the effective, global batch-size (i.e., scaling-out) \cite{b8}.
While the gradients computed over larger-batches are more representative of the true gradients (compared to full gradient descent) \cite{b5, b15}, it results in lower model quality over test/unseen data compared to training over a smaller batch-size.
In this paper, we address these trade-offs to propagate large-batch training in FL.
At intra-device level, we develop a parallel performance model to estimate the compute cost of training DNNs at different batches for the given compute hardware, thus maximizing efficiency.
We then cover prior art to improve statistical performance in large-batch training, and briefly introduce the notion of `gradient mapping' using a teacher model to attain convergence quality over large-batches akin to small-batch training.
	
	\BlankLine
	\BlankLine
	\section{Background and Related work}\label{sec:relatedwork}

In this section, we explore the trade-offs between parallel and statistical performance with different distributed training strategies, and explain how some techniques alleviate the generalization gap associated with large-batch training.

\subsection{Parallel and Statistical Performance Trade-Offs in FL}

FL may be centralized where clients updates communicate to a centralized parameter server or it may be decentralized (e.g., P2P, ring, etc.) \cite{b27} where updates are exchanged via collective communication \cite{b3}.
Federated algorithms may use a low-frequency, high-volume communication strategy \cite{b27} or a high-frequency, low-volume approach \cite{b28}.
\textit{Federated averaging} (\textit{FA}) \cite{b0} corresponds to the former, where updates are aggregated after certain training steps or epochs.
Here, each client `$c$' of $C$ devices updates parameters $w$ at iteration `$i$' over loss function $\mathcal{L}(\cdot)$ and samples $b^{(c)}$ from 
distribution $\mathcal{B}^{(c)}$ by computing local rounds $h$ after $H$ global communication rounds.
From Equation~(\ref{eqn:sgdupdate}), global model is aggregated over local models $w_{(i) + h}$ after every $h$ local rounds.

\begin{subequations}
	\begin{equation}
		w_{(i) + h + 1}^{(c)} = w_{(i) + h}^{(c)} - \eta \odot (\frac{1}{b^{(c)}} \sum_{s \in \mathcal{B}^{(c)}_{(i)+h}} \nabla \mathcal{L}_{s}(w_{(i) + h}^{(c)}))
		\label{eqn:sgdupdate}
	\end{equation}
	\begin{equation}
		w_{(i)+1} := \frac{1}{C}\sum_{c=1}^{C} w_{(i) + H}^{(c)}
		\label{eqn:fabspmodelupdate}
	\end{equation}
\end{subequations}

\BlankLine
Conversely, high-frequency, high-volume bulk synchronous-parallel (BSP) training aggregates updates across all clients at the end of \textit{every} iteration.
In Equation~(\ref{eqn:fabspmodelupdate}), this corresponds to $H=1$ so the global batch-size used to compute updates over each step is $B = \sum_{c=1}^{C}b^{(c)}$.
Assuming the same local batch-size across clients, $B=|Cb|$, implying that BSP performs more work `for every iteration' than local-SGD training.

\begin{table}
	\caption{Convergence quality under BSP and FA communication models}
	\centering
	\begin{tabular}{|c|c|c|c|}
		\hline
		\bfseries Configuration & \bfseries Partition & \bfseries Comm. & \bfseries Test Accuracy \\
		\hline
		\multirow{4}{*}{ResNet50 CIFAR10} & \multirow{2}{*}{IID} & BSP & \textbf{82.54\%} \\
		\cline{3-4}
		& & FA & 81.01\% \\
		\cline{2-4}
		& \multirow{2}{*}{Non-IID} & BSP & \textbf{65.35\%} \\
		\cline{3-4}
		& & FA & 58.72\%\\
		\hline
		\multirow{4}{*}{VGG11 CIFAR100} & \multirow{2}{*}{IID} & BSP & \textbf{83.91\%} \\
		\cline{3-4}
		& & FA & 74.5\% \\
		\cline{2-4}
		& \multirow{2}{*}{Non-IID} & BSP & \textbf{71.01\%} \\
		\cline{3-4}
		& & FA & 32.86\%\\
		\hline
		\multirow{4}{*}{AlexNet CalTech101} & \multirow{2}{*}{IID} & BSP & \textbf{67.1\%} \\
		\cline{3-4}
		& & FA & 62.67\% \\
		\cline{2-4}
		& \multirow{2}{*}{Non-IID} & BSP & \textbf{66.6\%} \\
		\cline{3-4}
		& & FA & 35.2\%\\
		\hline
	\end{tabular}
	\label{table:bspfedavgAccs}
\end{table}

Although federated algorithms have lower communication cost than BSP, the latter tends to achieve better statistical performance for a given training routine and data distribution.
We see this in Table (\ref{table:bspfedavgAccs}) where models trained over i.i.d. and non-i.i.d. datasets under BSP attain higher accuracy than federated averaging; non-i.i.d.  datasets are partitioned with 1, 10 and 10 labels per-client across a cluster of 10 devices over CIFAR10, CIFAR100 and CalTech101 datasets.
Updates are aggregated at every iteration in BSP, and once every training epoch in FA.
\textit{At the cost of higher synchronization overhead, BSP thus achieves better statistical performance than FA}.

\begin{table}
	\caption{Convergence in FA with different synchronization frequencies}
	\centering
	\begin{tabular}{|c|c|c|c|}
		\hline
		\bfseries Configuration & \bfseries Partition & \bfseries FA steps & \bfseries Test Accuracy \\
		\hline
		\multirow{4}{*}{ResNet50 CIFAR10} & \multirow{2}{*}{IID} & 250 & 78.07\% \\
		\cline{3-4}
		& & 1000 & 76.77\% \\
		\cline{2-4}
		& \multirow{2}{*}{Non-IID} & 250 & 67.31\% \\
		\cline{3-4}
		& & 1000 & 66.26\% \\
		\hline
		\multirow{4}{*}{VGG11 CIFAR100} & \multirow{2}{*}{IID} & 250 & 75.96\% \\
		\cline{3-4}
		& & 1000 & 70.79\% \\
		\cline{2-4}
		& \multirow{2}{*}{Non-IID} & 250 & 31.24\% \\
		\cline{3-4}
		& & 1000 & 12.0\% \\
		\hline
		\multirow{4}{*}{AlexNet CalTech101} & \multirow{2}{*}{IID} & 250 & 61.14\% \\
		\cline{3-4}
		& & 1000 & 59.62\% \\
		\cline{2-4}
		& \multirow{2}{*}{Non-IID} & 250 & 23.8\% \\
		\cline{3-4}
		& & 1000 & 24.1\% \\
		\hline
	\end{tabular}
	\label{table:fedavgStepsConfAccs}
\end{table}

In FA, the synchronization frequency is a key training parameter that determines overall speedup and convergence quality.
Table~(\ref{table:fedavgStepsConfAccs}) shows different models running i.i.d. and non-i.i.d. versions of various datasets in FA such that model is aggregated after every 250 and 1000 iterations respectively.
We observe that the higher the synchronization frequency (i.e., smaller the FA steps), the better is the test accuracy.
However, frequent synchronization incurs additional communication overhead, which may affect the cumulative training time.

\subsection{Parallel-Scaling Considerations}\label{subsec:parallelscaleFL}

Training can either be scaled-up (intra-device) or scaled-out (inter-device).
The synchronization frequency in distributed settings affects the overall iteration time and consequently, parallel performance.
During communication phase, cost is influenced by number of clients, model-size, device topology and collective implementation.
For e.g., the bandwidth cost rises with the cluster-size in centralized star-topology, while decentralized ring-allreduce is bandwidth-optimal \cite{b3}.
The process of forward and backpropagation at any step `$i$' is identical from a computational standpoint: taking time $t_{c}$ to compute model loss and gradients, $t_{mov}$ for data-batching and movement overhead (e.g., between system memory and CPU, system memory and GPU, across memory hierarchies, etc.), and if applicable, communication overhead (Equation~(\ref{eqn:itrcostFLBSP})).

\begin{equation}
	t_{step}^{(i)} = t_{c}^{(i)} + t_{mov}^{(i)} + ((i + H)\: \% H == 0\:?\:1: 0)\cdot t_{sync}^{(i)}
	\label{eqn:itrcostFLBSP}
\end{equation}

\BlankLine
In FL where models are synchronized after every $h \in [H]$ local steps across clients, so communication cost is zero in local-training phase and $t_{sync}$ during synchronization phase.
Thus, $|b|$ samples are used to train model on each client during local phase, and updates across $|Cb|$ samples are aggregated over the communication phase.

\begin{figure}
	\centering 
	\subfloat[ResNet50 on CIFAR10]{\includegraphics[width=0.25\textwidth]{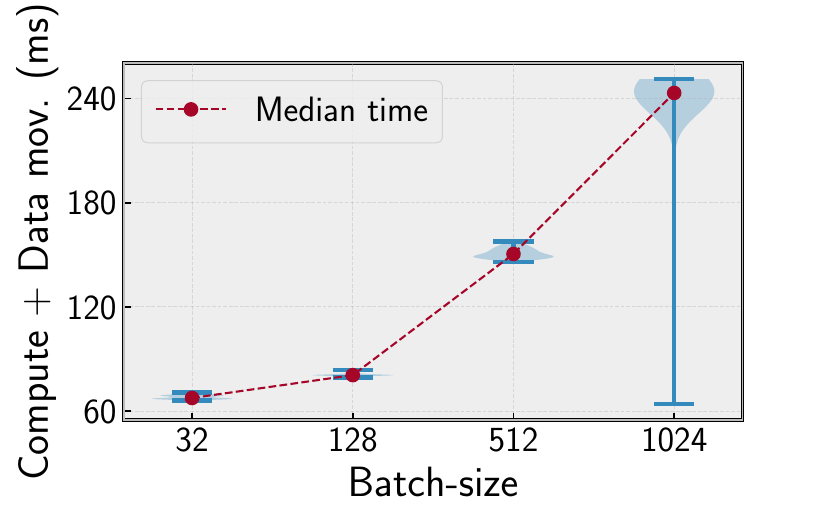}
	\label{res50computedatamov}}
	\subfloat[VGG11 on CIFAR100]{\includegraphics[width=0.25\textwidth]{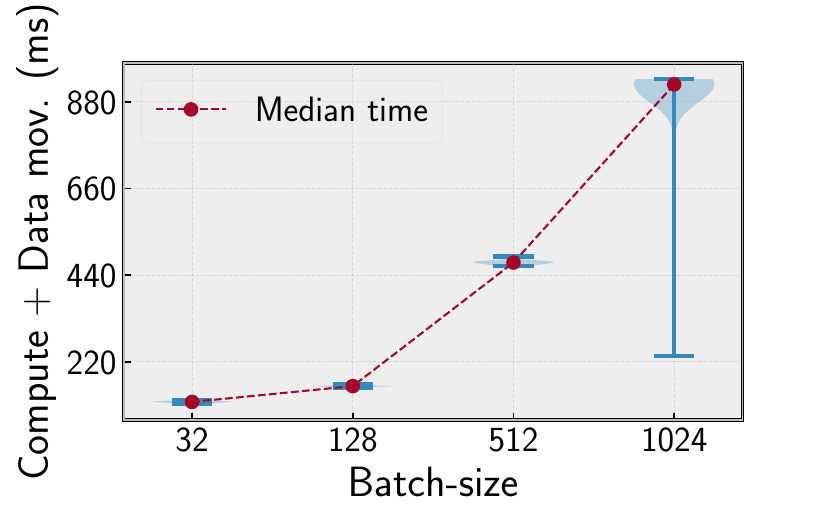}
	\label{vggcomputedatamov}}
	\hspace{0.01cm}
	\subfloat[AlexNet on CalTech101]{\includegraphics[width=0.25\textwidth]{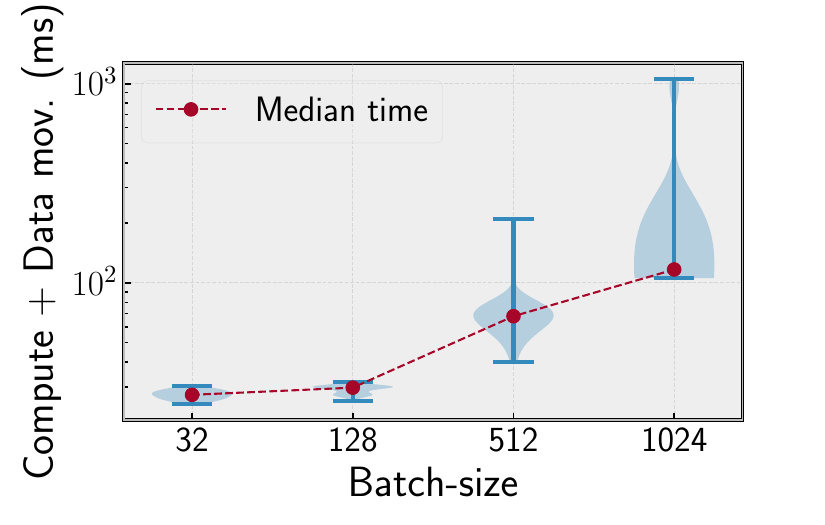}
	\label{alexcomputedatamov}}
	\subfloat[ViT on CalTech256]{\includegraphics[width=0.25\textwidth]{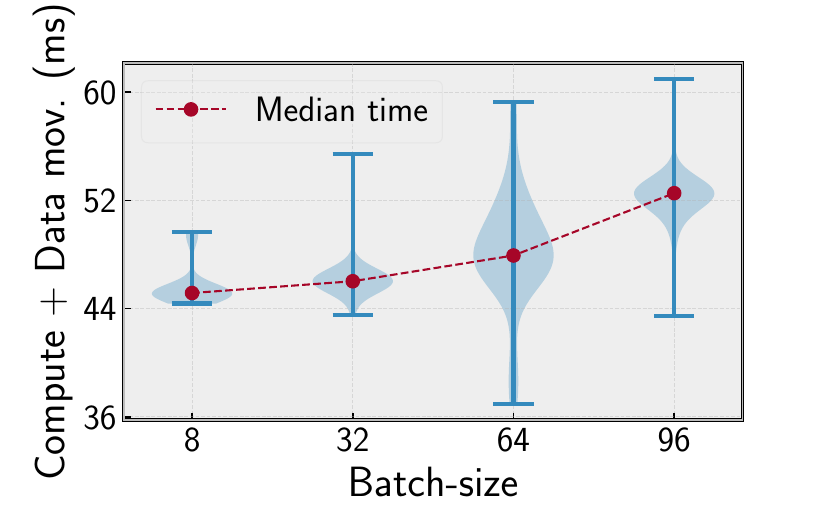}
	\label{vitcomputedatamov}}
	\caption{Computation time and data-movement overhead rises as batch-size increases.
	The cumulative overhead rises sharply in some cases (ResNet50 and VGG11), and gradually in others (AlexNet and ViT), thus depends on the performance characteristics of a model and dataset configuration.}	
	\label{fig:totalcomputedatamovTime}
\end{figure}

In BSP, training is \emph{scaled-out} by keeping device batch-size fixed and adding more devices (i.e., weak-scaling).
For $H=1$ in Equation~(\ref{eqn:itrcostFLBSP}), updates are aggregated with communication overhead $t_{sync}$ at every step.
By incurring this cost, updates are effectively computed and applied over $|Cb|$ samples on every step (thus performing more work per-iteration by essentially using larger batches).
On the other hand, training is \emph{scaled-up} by increasing the batch-size on a device.
Regardless of horizontal or vertical scaling, $t_{c}$ and $t_{mov}$ depend on a device's training sample-size $|b|$.
Both compute cost and data-movement overhead increases as device batch-size increases, as seen from their cumulative overhead in Figure~(\ref{fig:totalcomputedatamovTime}).
With larger batches, more data needs to be moved from disk/memory to devices' compute cores, while computation cost rises from calculating additional activation maps in the forward pass.
ViT uses smaller batches compared to other DNNs since the size of vision transformer model grows proportionally to batch-size, thus bounded by limited device memory.

\begin{figure}
	\centering 
	\subfloat[ResNet50]{\includegraphics[width=0.25\textwidth]{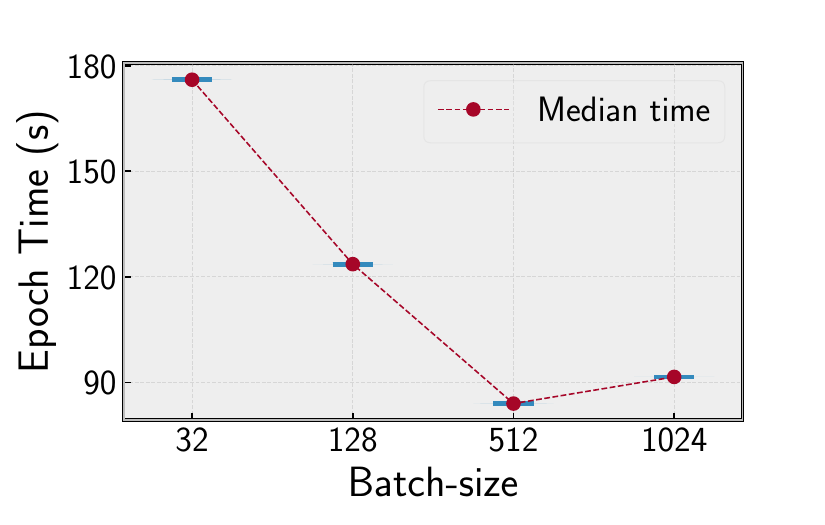}
	\label{res50perepochtime}}
	\subfloat[VGG11]{\includegraphics[width=0.25\textwidth]{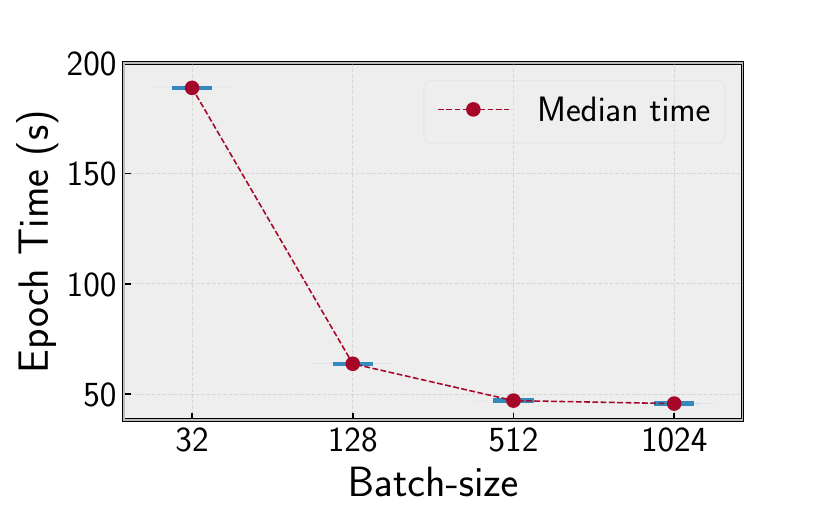}
	\label{vggperepochtime}}
	\hspace{0.01cm}
	\subfloat[AlexNet]{\includegraphics[width=0.25\textwidth]{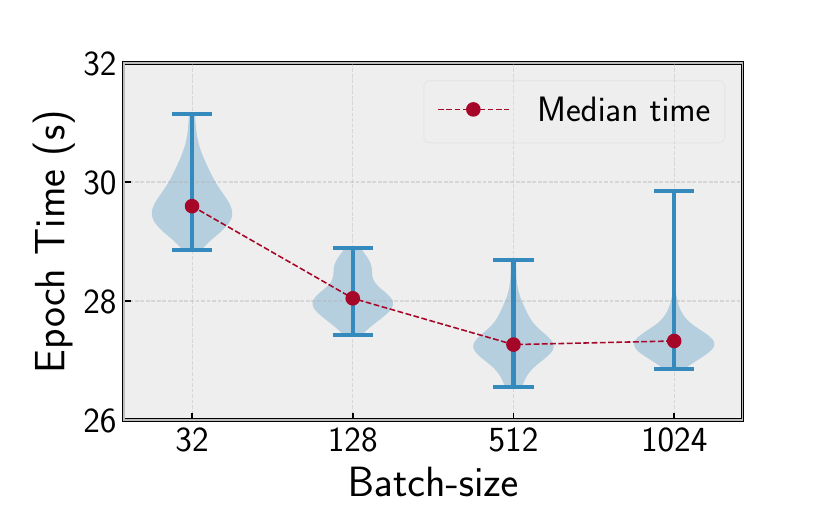}
	\label{alexperepochtime}}
	\subfloat[ViT]{\includegraphics[width=0.25\textwidth]{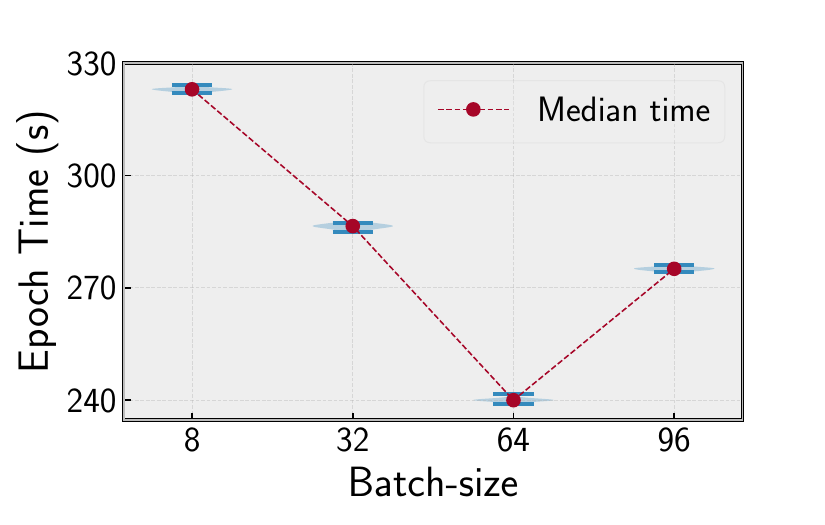}
	\label{vitperepochtime}}
	\caption{Distribution of time taken to complete one training epoch over different batches.
	Training performance improves initially as data-movement overhead is more costly than computation cost, and total iterations decreases as $B$ increases.
	However, compute cost outweighs as batch-size increases and performance eventually saturates.}	
	\label{fig:perEpochTime}
\end{figure}

\textit{In spite of higher $t_{c}$ and $t_{mov}$, larger batches tend to offer faster training performance}.
A model training over a dataset of size $D$ with batch-size $B$ takes iterations $I$ = $\frac{D}{B}$ to complete a single epoch; training for $E$ epochs thus takes $\frac{ED}{B}$ iterations.
So, large batches (i.e., $B$) take fewer iterations to run over training data.
The number of iterations directly affects the overall training time $T$ = $I \cdot t_{step}$ \cite{b15}.
Figure~(\ref{fig:perEpochTime}) shows the distribution and median time to complete a single epoch across various DNNs.
As $B$ increases, per-epoch time decreases as well, until batches get excessively large.
\textit{Extremely large-batches have diminishing returns, as it cannot leverage intra-device parallelism any further and so compute cost can rise significantly for even marginal increase in client's batch-size}.
We observe a higher epoch time at batch-size 1024 in ResNet50 and ViT, while the performance of VGG11 and AlexNet saturates at batch-size 512 over NVIDIA's K80 GPU.


\subsection{Improving Convergence in Large-Batch Training}

\begin{figure*}
\centering
\subfloat[ResNet50]{\includegraphics[width=0.25\textwidth]{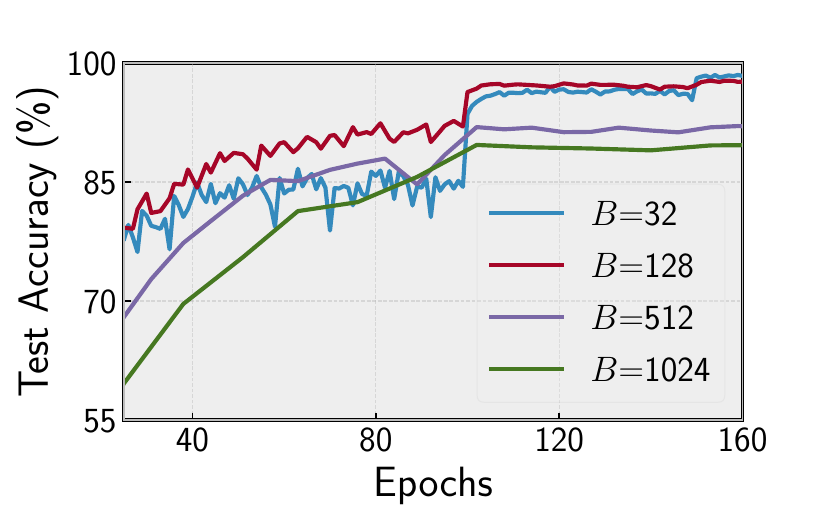}
\label{fig:res50lbtestacc}}
\subfloat[VGG11]{\includegraphics[width=0.25\textwidth]{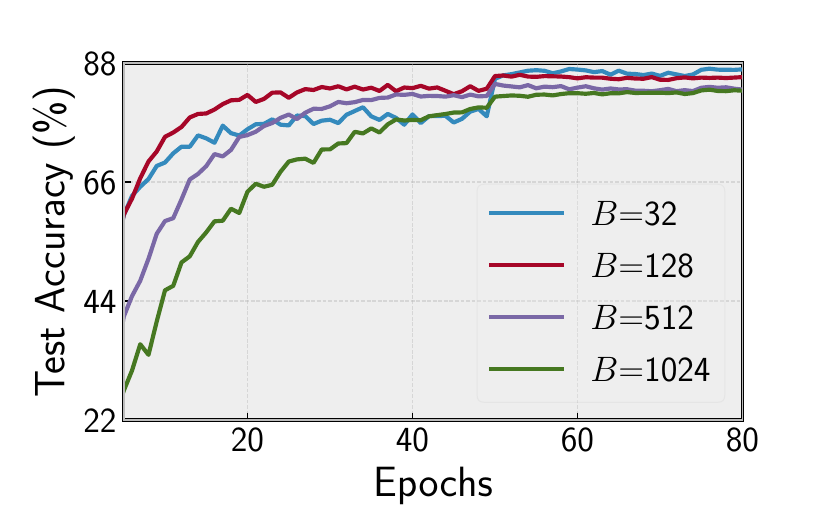}
\label{fig:vgg11lbtestacc}}
\subfloat[AlexNet]{\includegraphics[width=0.25\textwidth]{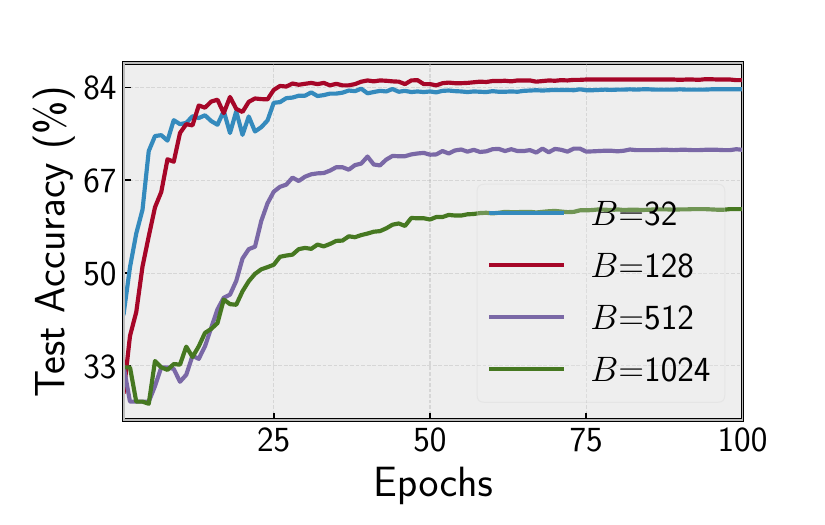}
\label{fig:alexlbtestacc}}
\caption{Test accuracy is worse over larger batch-sizes due to convergence over sharper minima.
When $B$ increases, generalization improves as models converge over flatter minima.
The generalization gap with large $B$ corresponds to a larger Hessian spectrum associated with sharp minima.} \label{fig:largebatchConvergence}
\end{figure*}

Although large-batches improve the parallel performance, it also results in worse statistical performance or generalization gap over small-batch training \cite{b9, b31}.
Large batches tend to converge to sharp minima, while small-batches explore flatter minima \cite{b11}.
\cite{b12} attributes this generalization gap with large-batches to a larger Hessian spectrum (commonly associated with sharp minima).
We observed similar accuracy degradation in test performance of ResNet50, VGG11 and AlexNet in Figure~(\ref{fig:largebatchConvergence}) where larger batches like 512 and 1024 achieve lower test accuracy than smaller batches.

Prior art mitigates this issue with techniques like learning-rate warmup, batch and layer normalization, adaptive learning-rates and gradient clipping.
To preserve SGD update (Equation~(\ref{eqn:sgdupdate})), \cite{b6} scaled learning-rate proportional to the batch-size and trained ResNet-50 on ImageNet with a batch-size of 8000.
\cite{b33} observed the optimal batch-size that maximized test performance to be proportional to learning-rate and size of training dataset.
\cite{b20} reduced total training iterations by using a larger learning-rate and proportionally scaling up its batch-size.
Post-local SGD \cite{b21} improved large-batch generalization by using BSP in the initial stages, followed by local-SGD.
\cite{b7} injected additional noise in the gradients to improve large-batch generalization.
Rather than inserting fixed Gaussian noise, \cite{b7} added perturbations in the form of diagonal Fisher noise.
\cite{b10} performed extrapolation to smoothen optimization trajectory and avoid sharp minima.
Here, gradients are computed at an extrapolated point, which is different from the current model state over which the SGD update is performed.
LARS or layer-wise adaptive-rate scaling adjusts learning-rate on each layer based on the magnitude of the gradients proportional to its weights \cite{b17, b4, b18}.
Inspired by LARS, LAMB \cite{b19} performed adaptive large-batch training with Adam optimizer and trained BERT in just over an hour.
TVLARS \cite{b32} replaced learning-rate warmup of LARS with a sigmoid-type function in the early training phase.
With elastic training, \cite{b5, b15} tune the global batch-size based on gradient noise.
\cite{b16} looks at adaptive batch-sizes through the lens of second-order Hessian information and uses larger batches if the eigenvalue decays by a certain factor over training.

In FL, a device's resource availability and data distribution may vary over time.
Thus, it is crucial to balance parallel and statistical performance aspect of large-batching training at both device-level and cluster-wide level for efficient FL.
	
	\BlankLine
	\BlankLine
	\section{Large-Batch Training}

To accelerate DNN training with large-batches while attaining high convergence quality across federated clients, we introduce a memory estimation model to predict the largest, usable batch-size given a client's computational resources.
We then propose a parallel performance model to infer the batch-size that achieves maximum training speedup.
Last, we give a broad overview of a statistical performance model that intelligently adds noise to large-batch updates to attain small-batch like generalizability.

\subsection{Estimating Batch-size Bounds}

\begin{figure}
\centering
\subfloat[$M_{act}$ (log-scale)]{\includegraphics[width=0.25\textwidth]{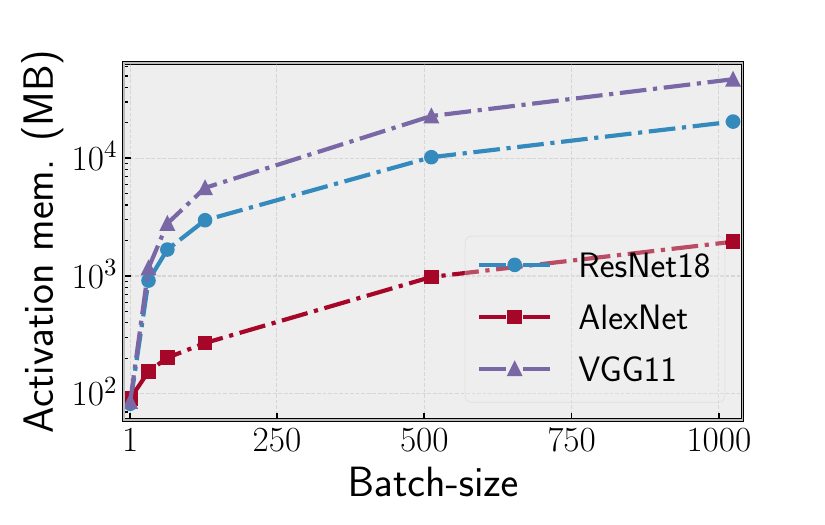}
\label{fig:actMem}}
\subfloat[Predicted vs. Actual $M_{batch}$]{\includegraphics[width=0.25\textwidth]{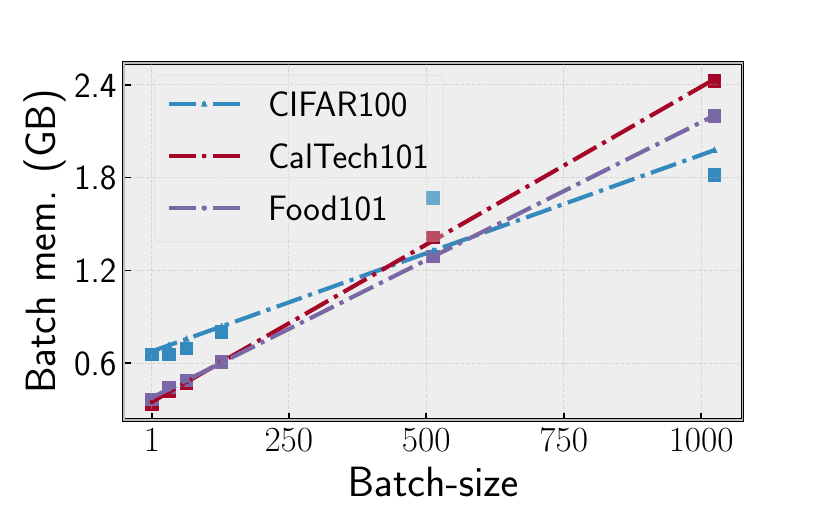}
\label{fig:dataloadMem}}
\caption{(a) Activation memory is proportional to the batch-size. (b) Batch memory is accurately predicted with a linear model.
Thus, given a model and dataset configuration, one can accurately predict the maximum permissible batch-size to run on a given device.} \label{fig:memoryEstimates}
\end{figure}

The maximum permissible batch-size on a client is limited by its available resources.
The total memory available on a device may limit the maximum batch-size that can be used without explicit checkpointing and recomputing intermediate model states \cite{b30}.
The memory required for DNN training depends on storing the model parameters ($M_{model}$) and gradients ($M_{grad}$), type and state of optimizer $M_{opt}$ (e.g., first or second-order), activation and batch memory(shown in Equation~(\ref{eqn:DNNMemory})).

\begin{equation}
	M_{total} = M_{model} + M_{grad} + M_{opt} + M_{act} + M_{batch}
	\label{eqn:DNNMemory}
\end{equation}

\BlankLine
\textit{The last two vary with training batch-size and thus, limit the maximum permissible batch-size}.
$M_{batch}$ corresponds to memory taken by the data-loading process to store and preprocess samples corresponding to the batch-size.
For fully-connected and convolution networks, $M_{act}$ is the total activation maps' memory computed on each sample in the forward pass.
Figure~(\ref{fig:actMem}) shows the activation memory of ResNet18, VGG11 and AlexNet on the log-scale and shows that $M_{act}$ rises linearly with the batch-size.
We also plot the actual batch-memory for different batch-sizes (i.e., 1, 32, 64, 128, 512 and 1024) as scatter points in Figure~(\ref{fig:dataloadMem}) and model the relationship between $M_{batch}$ and batch-size as a linear regression problem.
For CIFAR100, CalTech101 and Food101 datasets, the dashed lines show how the linear model predicts $M_{batch}$ for different batches.
CIFAR100 is predicted within 4.56-20.45\% error-rate via linear fitting, while CalTech101 has a prediction error of only 0.45-3.47\% and Food101 has an error range of 0.08-3.02\%.
\emph{By briefly running over extreme batch-sizes \cite{b15}, we can thus accurately predict the memory needed for training and determine maximum batch-size for a given hardware.}

\begin{figure}
	\centering 
	\subfloat[ResNet50]{\includegraphics[width=0.25\textwidth]{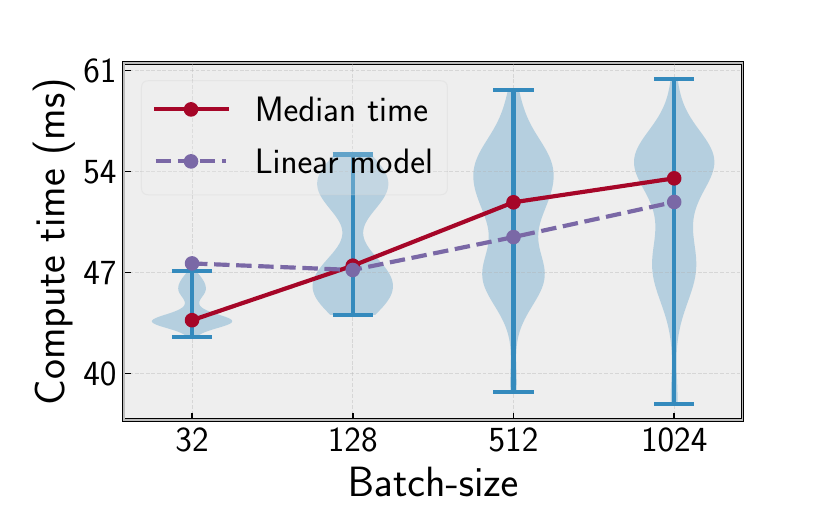}
	\label{res50ComputeTime}}
	\subfloat[VGG11]{\includegraphics[width=0.25\textwidth]{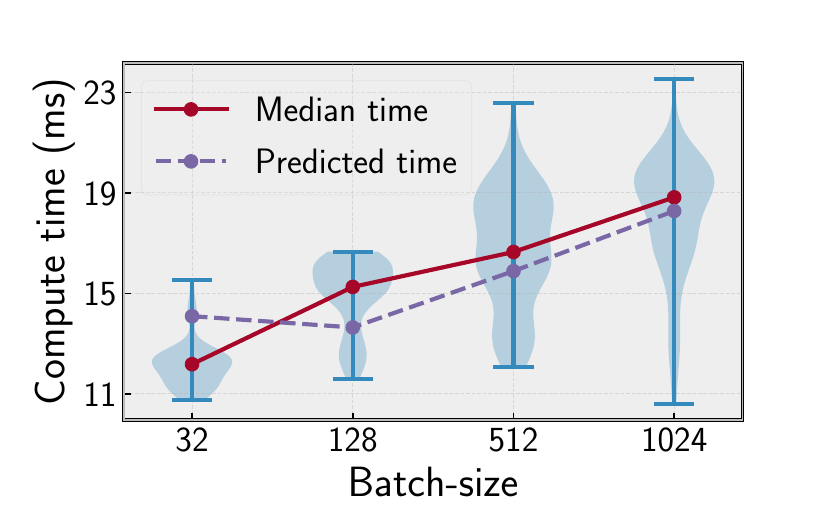}
	\label{vggComputeTime}}
	\hspace{0.01cm}
	\subfloat[AlexNet]{\includegraphics[width=0.25\textwidth]{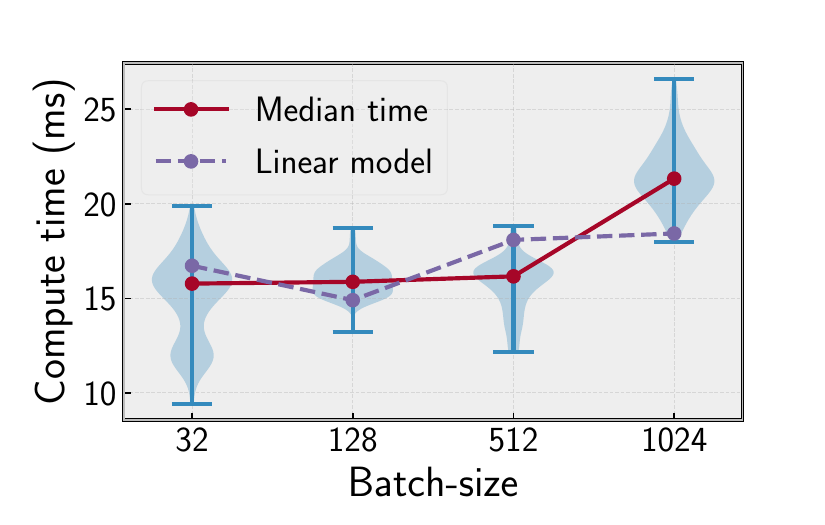}
	\label{alexComputeTime}}
	\subfloat[ViT]{\includegraphics[width=0.25\textwidth]{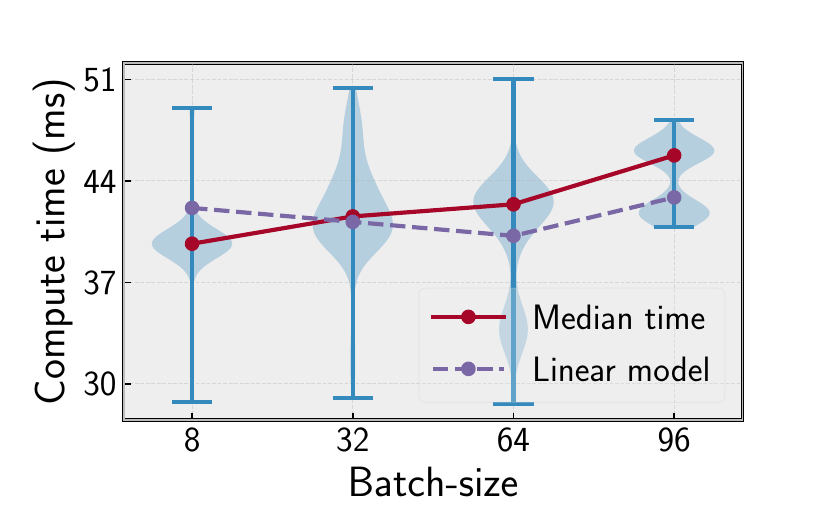}
	\label{vitComputeTime}}
	\caption{Compute time of different models over varying batch-sizes.
	Compared to the median compute time, the linear model  accurately predicts computation time within 9\%, 15.7\%, 13.6\% and 6.34\% error for ResNet50, VGG11, AlexNet and ViT respectively.}
	\label{fig:modelComputeTime}
\end{figure}

From \S\ref{subsec:parallelscaleFL}, we saw how the computation cost and data-movement overhead increased with larger batches.
Figure~(\ref{fig:modelComputeTime}) specifically shows the median and computation time distribution for different DNNs at varying batches.
The trend is that compute cost increases with larger batches, as more processing is needed to be done to compute additional activation maps in the feedforward phase.
Inspired by the parallel performance in \cite{b15}, we model compute time with respect to batch-size, i.e., $t_{c} \propto |b|$.
The dashed line shows the predicted time with a simplistic linear model fitted over the compute times logged for different batches (it does not look like a straight line as batch-sizes on the x-axis are not drawn to scale).
Linear interpolation predicts the compute time for a given batch-size with an error margin of 0.6-9\% in ResNet50, 2.88-15.7\% in VGG11, 5.98-13.6\% in AlexNet and 0.9-6.34\% in ViT.
\emph{From these results, we observe that a linear model is able to predict the computation time for a given batch-size within good accuracy.}
Accounting for even larger batches with synchronous data-parallel training can further be done by modeling communication cost for a given cluster and network configuration \cite{b15}.

With the aformentioned memory and computation cost prediction models, parallel performance can accurately be inferred at both inter- and intra-device level, thus enabling large-batch training by determining the largest permissible batch-size, and the optimal batch-size with the least compute cost.

\subsection{Improving Statistical Efficiency in Large-Batch FL}

\begin{figure}
\centering
\subfloat[ResNet50]{\includegraphics[width=0.25\textwidth]{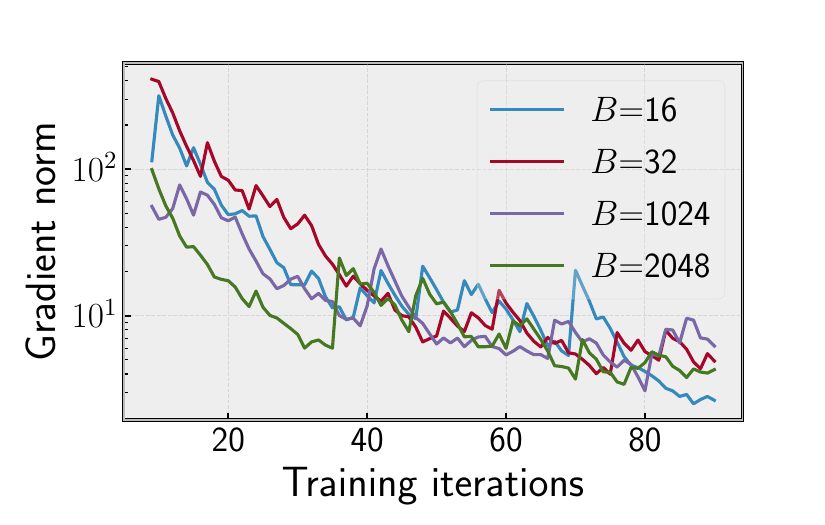}
\label{fig:res50GradNorm}}
\subfloat[VGG11]{\includegraphics[width=0.25\textwidth]{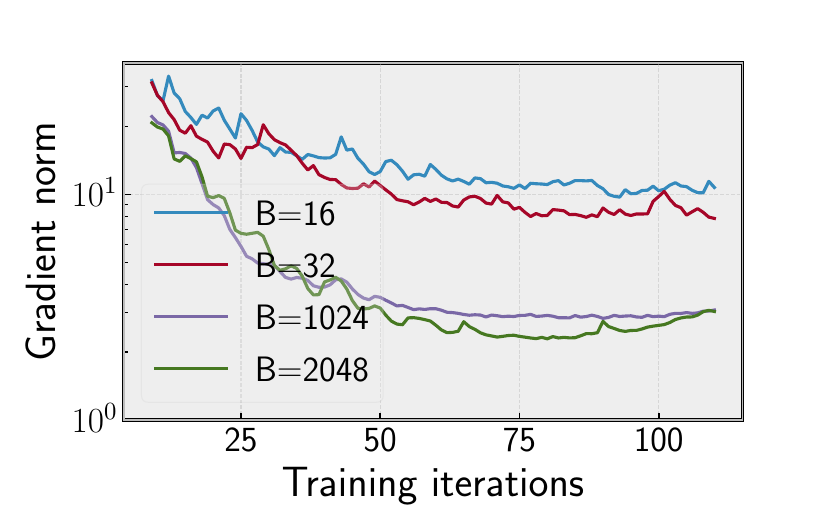}
\label{fig:vgg11GradNorm}}
\caption{Gradient norm in early training stages over different batches.
As $B$ increases, the signal-to-noise ratio of the gradients increases as well, thus updates get smaller and the gradient norm decreases.} \label{fig:trackGradNorms}
\end{figure}

Training DNNs with large-batches results in updates that are smooth and precise, i.e., closer to the true gradient had we used full-batch gradient descent \cite{b5, b15}.
However, this may lead to optimization towards sharp minima in the loss landscape, resulting in overfitting and poor generalization.
On the other hand, gradients calculated over smaller-batches tend to be noisy due to high variance \cite{b5}, which in turn provides additional regularization.
This additional noise in smaller-batches helps DNNs escape sharp minima and converge to flatter minima, thus improving test performance and even adding robustness to small perturbations.

\begin{equation}
	G^{(b_{small})} = G^{(b_{large})} + \gamma(b_{small}, b_{large})
	\label{eqn:smallgradlargegradsrelation}
\end{equation}

\BlankLine
We see noisy gradients in small-batch training (compared to large-batches) in early training stages of ResNet50 and VGG11 in Figure~(\ref{fig:trackGradNorms}).
We plot the cumulative gradient norm for different batches and observe that large-batch (i.e., $b_{large}$) updates tend to be smaller than small-batches ($b_{small}$), differing by gradient noise $\gamma$, shown as Equation~(\ref{eqn:smallgradlargegradsrelation}) \cite{b5, b15}.
The noise term here corresponds to the variance between the updates computed over small and large batches.
To improve the generalization gap in large-batch training, prior works have explored adding small perturbations or noise to gradient updates.
This allows for more exploration over the local minima across devices, and eventually converge to flatter minima.

\begin{subequations}
	\begin{equation}
		G^{(b_{large})}_{(i)} = \frac{1}{|B^{(b_{large})}|}\nabla \mathcal{L}(w_{(i)}, B^{(b_{large})})
		\label{eqn:largeGradsCompute}
	\end{equation}
	\begin{equation}
		\tilde{G}_{(i)} = \mathbf{\mathcal{A}}(G^{(b_{large})}_{(i)})
		\label{eqn:mappingLgradsSgrads}
	\end{equation}
	\begin{equation}
		w_{(i+1)} = w_{(i)} - \eta \odot \tilde{G}_{(i)}
		\label{eqn:updateSGDlargeGrads}
	\end{equation}
	\label{eqn:autoencoderGrad}
\end{subequations}

Rather than adding fixed noise at every iteration or compute the Fisher noise matrix \cite{b7}, we propose an alternate approach highlighted in Equation~(\ref{eqn:autoencoderGrad}).
Instead of using large-batch gradients $G^{(b_{large})}_{(i)}$ at iteration `$i$' to apply gradient descent update on model parameters, we use a pretrained encoder-like teacher model $\mathcal{A}(\cdot)$ that maps large-batch updates to an \emph{approximated} small-batch $\tilde{G}_{(i)}$.
The estimated gradient $\tilde{G}_{(i)}$ is then used to update model state for the next iteration (Equation~(\ref{eqn:updateSGDlargeGrads})).
The teacher model is ``briefly'' trained by deploying the student (i.e., the DNN to be trained with larger batches) with a small and a large batch-size respectively ($B^{(b_{small})}$ and $B^{(b_{large})}$).
Then, $\mathcal{A}(\cdot)$ is trained with gradients from $B^{(b_{large})}$ as input and generating gradients corresponding to $B^{(b_{small})}$ as the output.
We refer to $\mathcal{A}$ as a teacher and original DNN as the student since we want to learn the mapping from small to large-batch updates from $\mathcal{A}$.
The primary overhead with this approach comes from training $\mathcal{A}$; so the cost will be low if the teacher is trained briefly for a few iterations/epochs, and vice versa.
\emph{We don't necessarily need the teacher model to be highly accurate in predicting small-batch gradients; it simply needs to be able to map the variance between large and small-batch updates within reasonable estimates}.
Throughout training, the scale and trajectory of gradients may vary, as attributed to the initial training phase \cite{b35} and critical periods \cite{b34}.
Updating and retraining the teacher model in such phases may be required for accurate small-batch estimates.
Prior works have explored using autoencoders for adaptive gradient compression \cite{b13} and privacy-preserving FL \cite{b14}, but not for large-batch training.

\BlankLine
\begin{subequations}
	\begin{equation}
		\mathcal{U}(X) = 
		\begin{cases}
			1.0 & \text{in critical training phases} \\
			X & \text{otherwise}
		\end{cases}
		\label{eqn:stepfn}
	\end{equation}
	\begin{equation}
		\tilde{G}_{(i)} = \mathcal{U}(X, G^{(b)}_{(i)})
		\label{eqn:gradientStepFn}
	\end{equation}		
	\begin{equation}
		\tilde{G}_{(i)} = 
		\begin{cases}
			G^{(b)}_{(i)} & \text{in critical training phases} \\
			X \odot G^{(b)}_{(i)} & \text{otherwise}
		\end{cases}
		\label{eqn:tildeGradStepFn}
	\end{equation}
\end{subequations}

\BlankLine
\textbf{Na\"{\i}ve approach.} We assess the efficacy of this technique by substituting $\mathcal{A}(\cdot)$ with a simplistic step function $\mathcal{U}(\cdot)$ in our preliminary evaluation.
From Equation~(\ref{eqn:stepfn}), the step function steps-up or increases the gradients by factor $X$ when training is not in critical regions, and steps-down to the original, computed gradients under sensitive training phases.
Replacing $\mathcal{A}(\cdot)$ in Equation~(\ref{eqn:mappingLgradsSgrads}) with $\mathcal{U}(\cdot)$, we model gradients computed over larger batches as illustrated in Equations~(\ref{eqn:gradientStepFn}) and (\ref{eqn:tildeGradStepFn}).
Thus, we use the original gradients computed over the chosen batch-size $b$ in sensitive regions, and scale up the gradients by $X$ as if they were computed from a smaller batch-size.
This is in line with Equation~(\ref{eqn:smallgradlargegradsrelation}), where we set $\gamma$ to 0 in critical phases and inject artificial noise equivalent to $(X -1)G^{(b_{large})}$ otherwise (as shown in Equation~(\ref{eqn:modelNoiseWithStepFnGrads})).

\begin{equation}
	G^{(b_{small})} \approx G^{(b_{large})} + (X-1)G^{(b_{large})} = X \odot G^{(b_{large})}
	\label{eqn:modelNoiseWithStepFnGrads}
\end{equation}

\BlankLine
Thus, it is important to detect sensitive training phases in order to test this strategy.
From \cite{b1}, critical phase can be detected from changes in gradient norm over the iterations, as shown by the gradient change metric $\triangle(G_{(i)}^{(b)}$ in Equation~(\ref{eqn:criticalregionSelSync}).

\begin{equation}
		\triangle(G_{(i)}^{(b)}) = \bigg| \frac{|G_{(i)}^{(b)}|^{2} - |G_{(i-1)}^{(b)}|^{2}}{|G_{(i-1)}|^{2}} \bigg|
		\label{eqn:criticalregionSelSync}
\end{equation}
\BlankLine

\begin{figure}
	\centering 
	\subfloat[ResNet50 convergence]{\includegraphics[width=0.25\textwidth]{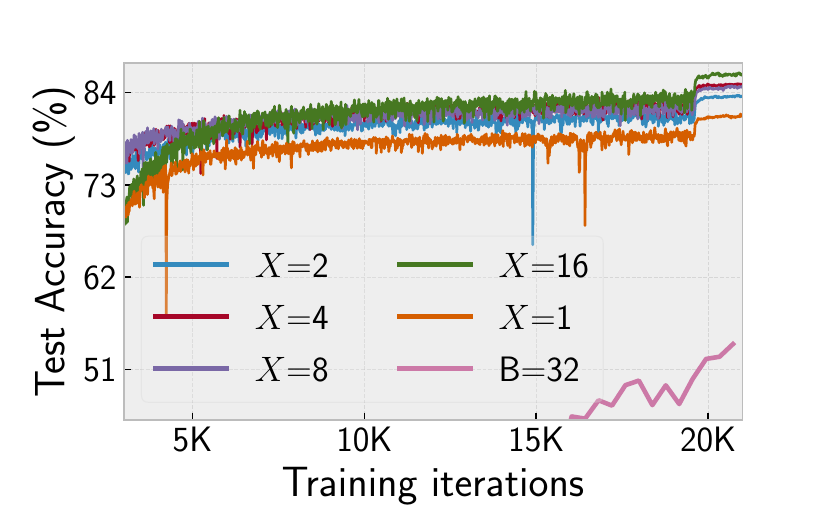}
	\label{fig:res50StepfnAcc}}
	\subfloat[ResNet50 KDE]{\includegraphics[width=0.25\textwidth]{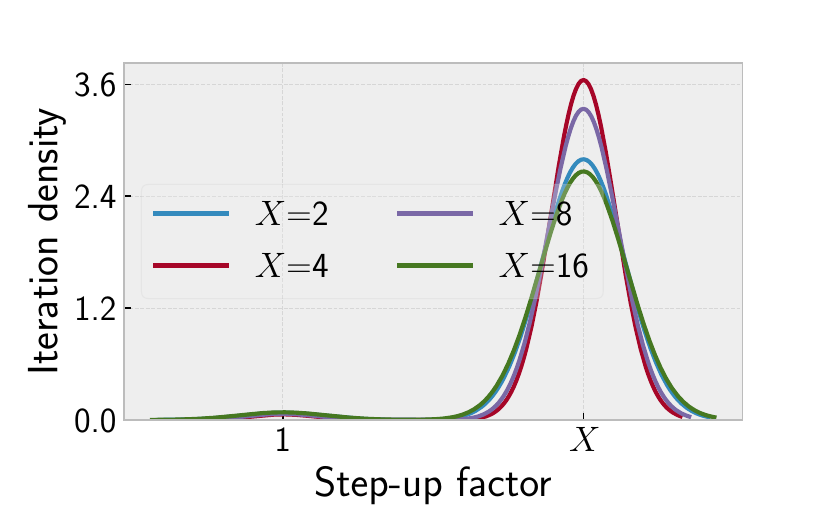}
	\label{rfig:es50StepfnItrKDE}}
	\hspace{0.01cm}
	\subfloat[VGG11 convergence]{\includegraphics[width=0.25\textwidth]{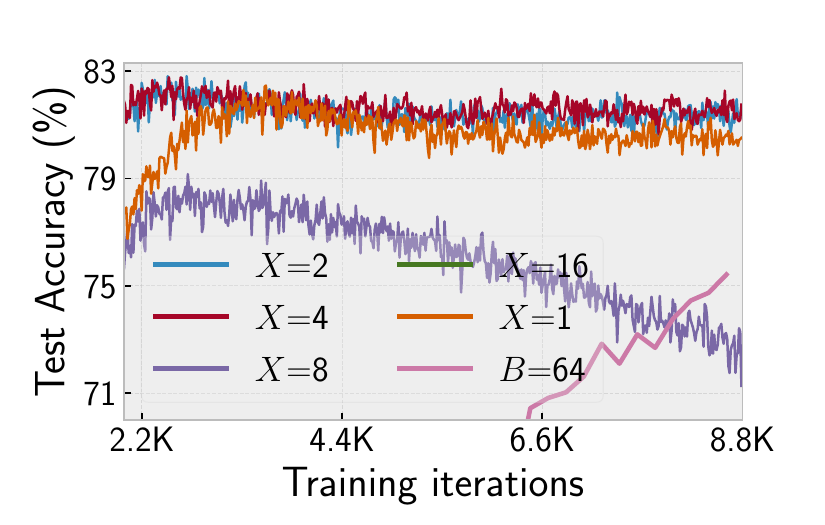}
	\label{fig:vgg11StepfnAcc}}
	\subfloat[VGG11 KDE]{\includegraphics[width=0.25\textwidth]{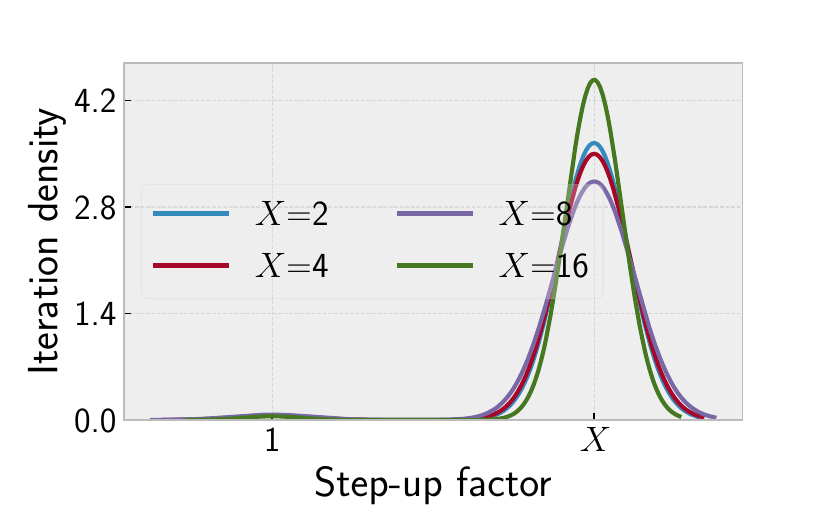}
	\label{fig:vgg11StepfnItrKDE}}
	\caption{Convergence quality and iteration kernel density estimates (KDE) when scaling-up the step function for various $X$ values \emph{at gradient change threshold 0.5}.
	Model quality improves over the baseline large-batch training (i.e., $X$=1) in most cases, but degrades in others when $X$ becomes too large (e.g. VGG11 with $X$=8 and $X$=16).}	
	\label{fig:accuracyKDEStepFn}
\end{figure}

\BlankLine
We train ResNet50 and VGG11, each with batch-size 1024 and step-up the computed gradients by setting $X$ to 2$\times$, 4$\times$, 8$\times$ and 16$\times$ respectively
In sensitive regions, we step-down the gradients by setting the step function to 1$\times$ (from Equation~(\ref{eqn:tildeGradStepFn})).
From \cite{b1}, we set a threshold of 0.5 on the rate of gradient change, so $\triangle(G_{(i)}^{(b)}) \geq 0.5$ steps-up the gradients to the chosen $X$, while updates are stepped down to 1$\times$ when $\triangle(G_{(i)}^{(b)}) < 0.5$.

We show the results in  Figure~(\ref{fig:accuracyKDEStepFn}) for different step-up factors along with $X$=1 (which corresponds to baseline training with a large batch-size 1024).
From Figure~(\ref{fig:res50StepfnAcc}), we see that the convergence quality of ResNet50 increases with $X$.
Here, $X$=16 attains higher accuracy than other step-up factors, and even more than the baseline configuration.
Thus, naively scaling up the gradient updates improves model quality and diminishes generalization gap even at larger batches.
Most of the iterations use scaling factor $X$ instead of 1$\times$ for $\triangle(G_{(i)}^{(b)}) \geq 0.5$, as seen from the kernel density estimates or KDEs of training iterations.
The KDE for $X$=1 is omitted since the original gradients are used to update model parameters and there is no additional step-up in this configuration (so only 1$\times$).
Thus, we see that even though majority of the updates scale-up the gradients, model quality improves even with higher scaling factors (i.e., $X$).
For reference with regard to small-batch training, Figure~(\ref{fig:res50StepfnAcc}) also plots the accuracy curve for batch-size 32, which shows that large-batch training is faster to converge in the same iterations, especially with gradient scaling. 
Compared to $B$=32, gradient scale-up with a large batch-size attained up to 32.33\% more accuracy.

However, convergence in VGG11 improved over the baseline only for $X$=2 and $X$=4, and degraded for larger $X$, as seen in Figure~(\ref{fig:vgg11StepfnAcc}).
The model loses considerable accuracy over $X$=8 (compared to $X$=1), and fails to improve beyond 10\% for $X$=16 (thus omitted from the figure).
From Figure~(\ref{fig:vgg11StepfnItrKDE}), most of the training iterations use scaling-factor $X$ (over 1$\times$) throughout the training phase, so larger stepping factors are detrimental to convergence quality in this case.

\begin{figure}
\centering
\subfloat[VGG11 convergence]{\includegraphics[width=0.25\textwidth]{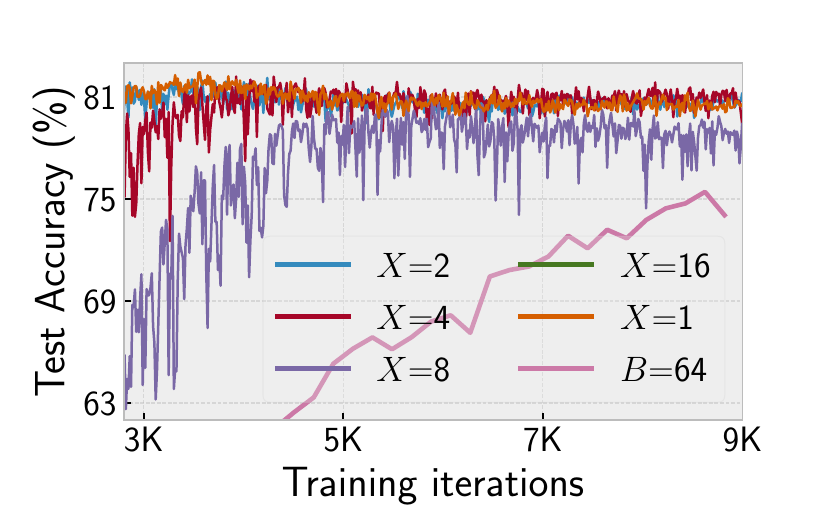}
\label{fig:vgg11StepfnAcc2}}
\subfloat[VGG11 KDE]{\includegraphics[width=0.25\textwidth]{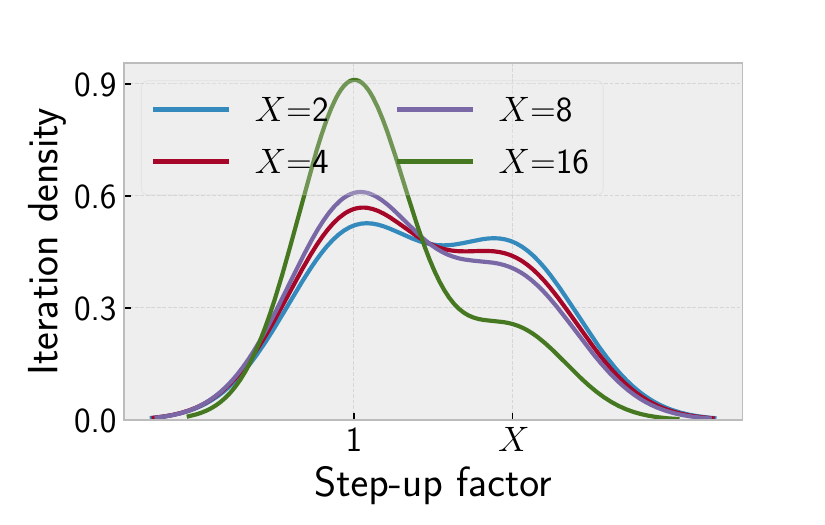}
\label{fig:vgg11StepfnItrKDE2}}
\caption{VGG11 convergence at various step-up factors \emph{with a higher gradient change threshold 0.8}.
Compared to the lower threshold (from Figure~(\ref{fig:accuracyKDEStepFn})), model quality improves at $X$=8 by stepping down to 1.0 for most iterations, as seen from the density plots.
However, VGG11 still failed to converge at $X$=16 as the scale-up factor was very large and diverged the model.} \label{fig:accuracyKDEStepFn2}
\end{figure}

To improve the convergence quality in VGG11, we vary the sensitivity of the $\mathcal{U}(\cdot)$ function by raising the gradient change threshold up to 0.8.
Thus, the step function is more censorious to model updates and uses the original gradients in majority of the iterations (i.e., 1$\times$).
The larger step-up factor $X$ is used only in highly sensitive iterations due to the higher $\triangle(G_{(i)}^{(b)})$ threshold.
Although $X$=8 configuration still underperformed $X \in$ [1,2,4], we see that compared to the lower threshold of 0.5 in Figure~(\ref{fig:vgg11StepfnAcc}), model accuracy significantly improved over the larger $X$=8 in Figure~(\ref{fig:vgg11StepfnAcc2}).
Even though $X$=16 still failed to achieve similar accuracy levels (thus, omitted from the results), it still attained higher accuracy of $\approx$ 30\% (compared to only 10\% for $X$=16 at threshold 0.5).
From Figure~(\ref{fig:vgg11StepfnItrKDE2}), the improvement in convergence quality can be attributed to intelligently switching between factors 1$\times$ and $X$ based on critical phase identification with a higher gradient change threshold.
Configurations $X \in$ [2,4,8] with the larger batch-size 1024 outperformed small-batch training with $B$=64 by attaining better accuracy in the same number of iterations.

\textit{\textbf{Summary}: With the aforementioned naive approach using a simple, static step-up function, some models and configurations perform better than baseline large-batch training, while others degrade a model's statistical performance (test loss or accuracy).
We observed the latter for larger $X$ in VGG11 model, and the former for all $X$ values in ResNet50 and smaller $X$ in VGG11.
Intuitively switching between the original and scaled updates based on critical phase detection is crucial for model convergence, where such sensitive periods are successfully identified with the gradient change metric.
However, as seen from the above results, the ideal threshold can vary with model and configuration, and thus requires carefully choosing the optimal value for this threshold.
We conjecture that replacing the stepping function $\mathcal{U}(\cdot)$ with a knowledge distilling teacher model $\mathcal{A}(\cdot)$ that maps large-batch gradients to their small-batch equivalent would improve the quality of estimated gradients even further, thus providing training speedup of large-batches as well as robust generalization commonly associated with small-batches.}

\BlankLine
\textbf{Gradient compression and large-batch training}: Frequent communication can become the performance bottleneck in synchronous training, especially when the model-size and cluster of federated devices is large.
Gradient compression techniques like sparsification, quantization or low-rank approximations \cite{b39} reduce the communication volume, thus accelerating the training process \cite{b28}.
In Equation~(\ref{eqn:compressoreqn}), a compression operator $\mathcal{C}(\cdot)$ reduces gradients $G^{(b)}$ computed over batch-size $b$ to compressed update $G^{(*)}$.

\begin{subequations}
	\begin{equation}
		G^{(*)} = \mathcal{C}(G^{(b)})
		\label{eqn:compressoreqn}
	\end{equation}
	\begin{equation}
		G^{(b)} = G^{(*)} + (G^{(b)} - G^{(*)})
		\label{eqn:smallBlargeBgradaccord}
	\end{equation}
	\begin{equation}
		G^{(b)} = G^{(*)} + \phi(b)
		\label{eqn:sparselargedensesmallG}
	\end{equation}
\end{subequations}

\BlankLine
Prior work \cite{b36} has explored the correlation between batch-sizes and compression during critical regimes by assuming the stochastic gradient $G^{(b)}$ to be composed of a sparse mean $G^{(*)}$ and a dense noise (i.e., $G^{(b)} - G^{(*)}$) for models trained with sparsity inducing norms, shown in Equation~(\ref{eqn:smallBlargeBgradaccord}).
Gradients computed over large-batches correspond to highly compressed updates, while small-batch training corresponds to weakly compressed gradients.
The backpropagated gradients $G^{(b)}$ can thus be decomposed into sparse, compressed updates $G^{(*)}$ and dense noise $\phi(b)$.
The relation between batch-size and gradient compression is further corroborated by the analogy between Equations~(\ref{eqn:smallgradlargegradsrelation}) and (\ref{eqn:sparselargedensesmallG}).
Thus, using large-batches in federated learning makes DNN training more robust to high degree of compression and provides considerable communication savings, resulting in good overall speedup.

\BlankLine
\textbf{Heterogeneity under federated learning}: Heterogeneity is pervasive across edge, cloud and high-performance computing.
The devices in a federated cluster may have inherent variance in their respective computational capability, or partial resource availability from concurrent execution of multiple jobs.
Heterogeneity may also rise from the data volume processed by each device.
To handle compute heterogeneity, \cite{b37, b38} proposed variable and dynamic-batching where each device is allocated a batch-size proportional to its compute capability.
To address data volume imbalance, prior works have explored weighted aggregation by scaling updates proportional to the number of samples processed by each device \cite{b0, b22, b37}.
In both resource and data volume heterogeneity, batches on some devices can become extremely small, and excessively large on others.
Using a teacher model as described in Equation~(\ref{eqn:autoencoderGrad}) may also help modulate the disproportional updates computed from variable/hetereogeneous training samples across devices.
	
	\BlankLine
	\BlankLine
	\section{Conclusion and Future Work}

It is crucial to consider both parallel and statistical efficiency in large-batch training for fast, scalable and accurate FL systems.
In this work, we first introduce a parallel performance model to estimate the optimal batch-size that minimizes training time, and determine the largest, permissible batch-size usable on a given hardware.
Using a larger batch-size for training can improve the execution performance of federated learning in two ways: one, by lowering cumulative iterations over the epochs and reducing the data-movement overhead, and second, by allowing higher compression factors in compression-enabled distributed training.

To improve the generalization gap with large-batches, we envision an encoder-based teacher-student model configuration for knowledge distillation where the teacher would output small-batch updates from the large-batch gradients generated by the student model.
We additionally show a simple step function (as a naive substitute to the encoder-based model) that either scales-up or scales-down the gradients by a fixed value under certain training periods can significantly improve the convergence quality over vanilla large-batch training.
Compared to small-batch settings, we also observed higher accuracy over the same set of training iterations.
In our preliminary evaluation, gradients scaled to 16$\times$ in less sensitive regions and 1$\times$ (i.e., the original updates) under critical phases attained up to 4.89\% better accuracy than the baseline (corresponding to large-batch training with $X$=1 at batch-size 1024) and 32.33\% higher accuracy than small-batch training (with batch-size 32) in ResNet50.
In VGG11, although the naive step function at 8$\times$ scale-up attains about 3.3\% lower accuracy than baseline, but still achieved 3.74\% more accuracy than small-batching training for the same iterations.

In future work, we intend to explore different architectures like transformer encoders and autoencoders for the proposed teacher model.
Since the objective of teacher-model is to learn the mapping between updates computed over a large and small batch-size, another challenge is to determine how large or small the chosen batches should be.
Based on prior work, we will study gradient noise to determine the optimal batches to be used.
Last, we will focus on using adaptive batch-sizes in conjunction with gradient compression in order to attain high model quality with minimal communication overhead.

\end{document}